# Enhancing Convolutional Neural Networks with Higher-Order Numerical Difference Methods

1st Qi Wang
Northeastern University
Boston, USA

2nd Zijun Gao
Northeastern University
Boston, USA

3rd Mingxiu Sui
University of Iowa
Iowa City, USA

4th Taiyuan Mei
Northeastern University
Boston, USA

5th Xiaohan Cheng
Northeastern University
Boston, USA

6th Iris Li *
New York University
New York, USA

*Abstract*—With the rise of deep learning technology in practical applications, Convolutional Neural Networks (CNNs) have been able to assist humans in solving many real-world problems. To enhance the performance of CNNs, numerous network architectures have been explored. Some of these architectures are designed based on the accumulated experience of researchers over time, while others are designed through neural architecture search methods. The improvements made to CNNs by the aforementioned methods are quite significant, but most of the improvement methods are limited in reality by model size and environmental constraints, making it difficult to fully realize the improved performance. In recent years, research has found that many CNN structures can be explained by the discretization of ordinary differential equations. This implies that we can design theoretically supported deep network structures using higher-order numerical difference methods. It should be noted that most of the previous CNN model structures are based on low-order numerical methods. Therefore, considering that the accuracy of linear multi-step numerical difference methods is higher than that of the forward Euler method, this paper proposes a stacking scheme based on the linear multi-step method. This scheme enhances the performance of ResNet without increasing the model size and compares it with the Runge-Kutta scheme. The experimental results show that the performance of the stacking scheme proposed in this paper is superior to existing stacking schemes (ResNet and HO-ResNet), and it has the capability to be extended to other types of neural networks.

## I. INTRODUCTION

Neural networks are mathematical models that simulate the way neurons operate in the brain. The first neural network structure to emerge was the feedforward neural network (FNN) [1], which typically consists of multiple layers of interconnected neurons. When an FNN receives input, the input layer nodes receive the input information and map it, then pass the mapped information through the connections between layers to the next layer of neurons. The entire training process of the FNN repeats the above steps between layers until the model converges. After training, we can use the trained FNN to predict test data. In the past, FNNs have been widely used to handle various problems involving formatted data and have achieved great success.

However, with the development of network technology, the information carriers we face are no longer limited to numbers. Text, images, and even videos are often the data sources encountered in practice. For such data, FNNs cannot process and learn effectively, so we prefer neural network structures that can directly handle these unstructured data sources. Taking images as an example, the most common image resolution we encounter is 19201080. If we use a 19201080 image as the learning data for the neural network, then for the FNN, the number of neurons in the input layer would be as high as 2,073,600. If we also consider the number of neurons in each hidden layer, then the number of parameters for the entire FNN would be very large. Such a model with a huge number of parameters is very difficult for us to train or deploy. To solve this problem, CNN was proposed in 1998. CNNs differ from FNNs in that they can accept matrices as input. When using neural networks to process image problems, computers usually process the image into multiple channels of matrices and then input them into the CNN for learning [2]. The mapping of data by neurons in CNNs is different from that in FNNs. Convolutional kernels (filters) are introduced in CNNs, which are matrices of a fixed size. Convolutional kernels map the image information by multiplying and summing with the corresponding pixel positions to extract information from the image.

By introducing different convolutional kernels, each kernel learns different features, which allows the image to be reorganized from various aspects into a form of numerical information that computers can understand and further analyze to solve the problems we expect neural networks to solve.

Over the past few decades, thanks to the development of CNNs, various AI-driven machines have acquired capabilities similar to human eyes and have successfully performed common computer vision tasks such as medical image recognition in survival prediction [3], few-shot description[4], and scene segmentation [5]. Looking back at the development of CNNs, it is not difficult to find that the existing CNN structures are very different from the beginning. In terms of depth improvement, there are neural networks represented by ResNet, Inception, and DenseNet; in terms of channel information utilization improvement, there are neural networks represented by SENet and ShuffleNet; in terms of attention improvement, there are neural networks represented by CoAtNet and CBAM. Although the above improvement methods have improved the performance of CNNs, most of the improvement methods are based on inspiration or rely on long-term deep learning engineering experience. In most practical scenarios, engineers usually hope to obtain stable performance improvements by simply changing the network structure. Therefore, improvement methods based on inspiration and long-term work experience are not suitable for deep learning engineers to quickly improve CNNs and apply them to practical problems. Based on this, an improvement algorithm with sufficient theoretical support and interpretability is particularly needed.

In past research, some researchers have found that the forward propagation process of ResNet-like neural networks can be seen as the process of solving ODEs by the forward Euler difference method in numerical difference methods, and its stacking structure can be seen as the form of the forward Euler difference method. Generally speaking, for the same ordinary differential equation, higher-order numerical methods have higher solution accuracy. Therefore, this chapter first constructs a high-order numerical difference method and proves the superior performance of the proposed numerical difference method through numerical experiments. Subsequently, this chapter improves the existing ResNet Euler stacking scheme with the proposed Taylor multi-step method and verifies the superiority of the proposed method on datasets such as CIFAR 10/100 and SVHN. Finally, this chapter also discusses the impact of network depth on the improvement method, the computational efficiency of the improvement method, and the scalability of the improvement method on other ResNet-like networks

## II. BACKGROUND

In this paper, we analyze various numerical methods and present corresponding network structures. Initially, the shortcomings of the forward Euler method are discussed. Subsequently, the paper introduces numerical methods with smaller truncation errors, such as the Improved Euler Method $O(\tau^3)$, the 2nd-order Runge-Kutta Method $O(\tau^3)$, the 3rd-order Runge-Kutta Method $O(\tau^4)$, the 3rd-order Heun Method $O(\tau^4)$, and the 4th-order Runge-Kutta Method $O(\tau^5)$. Unlike the aforementioned improvement methods, this paper utilizes the Taylor expansion to construct a high-precision multi-step numerical difference method and analyzes its superiority over the aforementioned numerical methods through numerical experiments.

## III. METHOD

### A. Forward Euler Algorithm

The forward Euler method is a first-order numerical method for solving a given initial value ordinary differential equation. Assume that the function $\theta(t)$ is discretized and has mm discrete points. Therefore, the time step $\tau = t_{l+1} - t_l (0 \leq l \leq m - 1)$. The value of $\theta(t)$ at time $l+1$ can be written as $\theta(t_{l+1}) = \theta(t_l + \tau)$. Furthermore, using the Taylor series expansion, we can obtain $\theta(t_{l+1}) = \theta(t_l) + \tau \cdot \theta'(t_l) + O(\tau^2)$. Finally, we can derive the following forward Euler equation: $\theta(t_{l+1}) = \theta(t_l) + \tau \cdot \theta'(t_l)$. Subsequently, the literature [6] found that ResNet is related to the forward Euler equation, and each of its basic blocks can be written as $x_{i+1} = x_i + F(x_i)$,

In this paper, $i \in \mathbb{Z}^+$ represents the depth of the residual block; $x_i$ represents the input of the *i*-th residual block; $x_{i+1}$ represents the output of the *i+1*-th residual block, and the input of the *i+1*-th residual block. $F(x_i)$ in Equation is equivalent to $\tau \cdot \theta'(t_l)$, representing the residual mapping that the neural network needs to learn. The structure of the residual block is given to illustrate the relationship between the forward Euler equation and ResNet. Essentially, ResNet is a kind of residual network. We can understand it as a sub-network that, when stacked, can form a very deep network. Generally speaking, as the depth of the neural network increases, the more information it can obtain, and the richer the features it learns.

However, experiments have shown that as the network deepens, the optimization effect actually worsens, and the accuracy of the test data and training data actually decreases. This is because deepening the network can make the deep layers become an identity mapping layer, then the model degenerates into a shallow network, followed by the problems of gradient explosion and gradient disappearance. Therefore, to alleviate the problems of gradient disappearance and gradient explosion, another way is needed to let the model learn the identity mapping function. Because if you let some layers fit a potential identity mapping function directly, it is usually quite difficult. We can transform it into learning a residual function $F(x_i) = x_{i+1} - x_i$. As long as $F(x_i) = 0$,, it constitutes an identity mapping. Combining the experimental results, the method of residual learning has successfully allowed the deep network to learn the identity mapping, and ResNet is using the method of residual learning to effectively alleviate the problems of gradient disappearance and gradient explosion.

### B. Runge-Kutta Methods

For the given differential equation $\theta' = f(t, \theta)$. The forward Euler method for solving the differential equation is given by

$$\begin{cases} \theta(t_{l+1}) = \theta(t_l) + \tau g^*, \\ g^* = f(t_l, \theta(t_l)) \end{cases},$$

where $g^*$ represents the average gradient. Generally speaking, a more accurate average gradient can help numerical methods achieve better numerical solutions. Therefore, the function values at certain special points $(f(t_k, \theta(t_k), t_k \in [t_l, t_{l+1}])$ are commonly used to calculate and estimate a better gradient $g^*$, thereby improving the computational accuracy of the forward Euler method. In other words, such improved methods actually estimate the gradients at several points within the interval $[x_n, x_{n+1}]$ and take their weighted average to estimate the average gradient $g^*$. Through such improved methods, the forward Euler method can achieve higher computational accuracy. Based on this idea, the Improved Euler (IE) method, can be expressed as:

$$\begin{cases} k_1 = F(t_l, \theta(t_l)) \\ k_2 = F\left(t_l + \frac{\tau}{2}, \theta(t_l) + \frac{\tau k_1}{2}\right) \\ \theta(t_{l+1}) = \theta(t_l) + \frac{\tau}{2} \cdot (k_1 + k_2) + O(\tau^3) \end{cases}$$

The 2nd Runge-Kutta (RK2) method, can be expressed as:

$$\begin{cases} k_1 = F(t_l, \theta(t_l)), \\ k_2 = F\left(\frac{2t_l}{3}, \theta\left(\frac{2t_l}{3}\right)\right), \\ \theta(t_{l+1}) = \theta(t_l) + \frac{\tau}{4} \cdot (k_1 + 3k_2) + O(\tau^3)_\circ \end{cases}$$

The 3rd Runge-Kutta (RK3) method can be expressed as:

$$\begin{cases} k_1 = F(t_l, \theta(t_l)), \\ k_2' = F\left(t_l + \frac{\tau}{3}, \theta(t_l) + \frac{\tau}{3} k_1\right), \\ k_2 = F\left(t_l + \frac{2\tau}{3}, \theta(t_l) + \frac{2\tau k_2'}{3}\right), \\ \theta(t_{l+1}) = \theta(t_l) + \frac{\tau}{4} \cdot (k_1 + 3k_2) + O(\tau^4)_\circ \end{cases}$$

The 4th Runge-Kutta (RK4) method can be derived. The Improved Euler method can be expressed as:

$$\begin{cases} k_1 = F(t_l, \theta(t_l)), \\ k_2 = F\left(t_l + \frac{\tau}{2}, \theta(t_l) + \frac{\tau}{2} k_1\right), \\ k_3 = F(t_l + \tau, \theta(t_l) + 2\tau k_2 - \tau k_1), \\ \theta(t_{l+1}) = \theta(t_l) + \frac{\tau}{6} \cdot (k_1 + 4k_2 + k_3) + O(\tau^4)_\circ \end{cases}$$

*C. Taylor Multistep Method*

The forward Euler method is the simplest numerical method, and the results obtained by solving ODEs with it are usually of limited accuracy. Therefore, we typically use higher-order numerical methods to solve ordinary differential equations. This paper derives a multi-step high-order numerical difference method from the Taylor series and proposes an effective stacking scheme. The derivation of the multi-step numerical difference method is as follows:

$$\theta(t_{l+1}) = \theta(t_l + \tau)$$
$$= \theta(t_l) + \tau \cdot \theta'(t_l) + \frac{\tau^2}{2!} \cdot \theta''(t_l)$$
$$+ \frac{\tau^3}{3!} \cdot \theta'''(\xi_1)$$
$$\theta(t_{l-1}) = \theta(t_l - \tau)$$
$$= \theta(t_l) - 2\tau \cdot \theta'(t_l) + \frac{(2\tau)^2}{2!} \cdot \theta''(t_l)$$
$$- \frac{(2\tau)^3}{3!} \cdot \theta''''(\xi_3),$$

Where $\xi_1, \xi_2$, and $\xi_3$ are respectively within the time steps $(l, l + \tau), (l - \tau, l), (l - 2\tau, l)$. Subsequently, Equations can be written as:

$$\theta'(t_l) = \frac{\theta(t_{l+1}) - \theta(t_l)}{\tau} - \frac{\tau}{2!} \cdot \theta''(t_l)$$
$$- \frac{\tau^2}{3!} \cdot \theta'''(\xi_1),$$
$$\theta'(t_l) = \frac{\theta(t_l) - \theta(t_{l-1})}{\tau} + \frac{\tau}{2!} \cdot \theta''(t_l)$$
$$- \frac{\tau^2}{3!} \cdot \theta'''(\xi_2),$$
$$\theta'(t_l) = \frac{\theta(t_l) - \theta(t_{l-2})}{2\tau} + \tau \cdot \theta''(t_l)$$
$$- \frac{2\tau^2}{3} \cdot \theta'''(\xi_3).$$

To obtain the following equation:
$$\theta'(t_l) = \frac{2\theta(t_{l+1}) - 3\theta(t_l) + 2\theta(t_{l-1}) - \theta(t_{l-2})}{2\tau} + \delta,$$

Where $\delta = \tau^2(-\theta'''(\xi_1)/3! - 2\theta''''(\xi_3)/3 + \theta'''(\xi_2))/3!$. Finally, a high-order differential equation is obtained as follows:

$$\theta(t_{l+1}) = \frac{3}{2}\theta(t_l) - \theta(t_{l-1}) + \frac{1}{2}\theta(t_{l-2}) + \tau\theta'(t_l) + 2\tau\delta$$

We notice that the truncation errors of the forward Euler method and the proposed multi-step numerical difference method are $O(\tau^2)$ and $O(\tau^3)$, respectively. This means that the truncation error order of the constructed multi-step difference method is higher than that of the forward Euler method.

It is known that the numerical method proposed in this paper is superior to the forward Euler method. Inspired by the results of previous research [7-9], we designed a new residual connection method using the proposed Taylor multi-step method and applied it to build TM-ResNet to improve ResNet. In the literature [10], the activation function in the structure of PreActResNet does not appear before the residual connection. Such a design ensures the consistency of information transmission in the residual connection, thereby improving the model's performance, and thus PreActResNet is more in line with the definition of the forward Euler method.

Therefore, the TM-ResNet proposed in this paper is designed based on PreActResNet. It should be noted that there are three initial values in Equation. Therefore, as shown in Figure 1, we use three residual blocks to construct the Boot TM-block. This structure can help us obtain the three initial values required to start TM-ResNet.

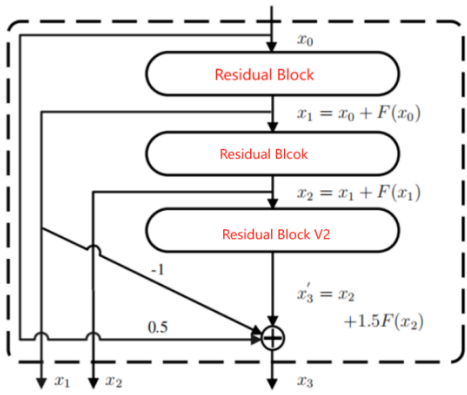

Figure 1. Boot T-Block Model Structure

## IV. EXPERIMENT

### A. Experiment Settings

*1) Dataset*

To verify the superiority of the proposed improved method, this chapter conducted experiments on four datasets.

1) CIFAR-10. The CIFAR-10 dataset consists of 60,000 32x32 color images in 10 categories, with 50,000 images for training and 10,000 images for testing.

2) CIFAR-100. The CIFAR-100 dataset has the same amount of data as the CIFAR-10 dataset, but the CIFAR-100 dataset divides images into 100 categories.

3) SVHN. SVHN is an image dataset, which is used to assist models in recognizing letters and numbers in natural scenes.

4) YouTube's face dataset. This dataset includes videos of most celebrities on YouTube. In our experiments, these videos have been processed into a facial keypoint dataset, which contains 5770 faces and their associated facial key points.

*2) Setting and Parameter*

To ensure a fair comparison between models, the same training parameters were used for all models being compared during the training process, and all experiments were conducted multiple times. In the image classification experiments and scene text recognition experiments, the training was set for 120 epochs with a batch size of 256.

To make the experiments more convincing, multiple models of different depths were constructed for each scheme and compared. When comparing the performance differences between models before and after the improvement for different network depths caused by different network architectures, the experiments ensured that the parameter volume of the networks involved in the comparison was approximately the same.

### B. Experiment Results Analysis

Table 1. Experiment Result

| Model | Parameter (CIFAR10) | Acc(CIFAR10) | Parameter (CIFAR100) | Acc(CIFAR100) |
|---|---|---|---|---|
| PreActResNet-18 | 11,171k | 91.54 | 11,217k | 73.28 |
| TM-ResNet-22 | 11,001k | 93.01 | 11,047k | 71.28 |
| PreActResNet-34 | 21,27k | 90.94 | 21,325k | 73.53 |
| TM-ResNet-36 | 21,035k | 93.44 | 21,081k | 74.97 |
| PreActResNet-50 | 23,509k | 92.79 | 23,693k | 74.77 |
| TM-ResNet-44 | 23,511k | 93.81 | 23,696k | 74.88 |
| PreActResNet-101 | 42,501k | 93.53 | 42,685k | 75.35 |
| TM-ResNet-86 | 42,497k | 93.42 | 42,682k | 76.50 |
| PreActResNet-152 | 58,144k | 93.68 | 58,329k | 75.30 |
| TM-ResNet-137 | 58,141k | **94.14** | 58,325 | **76.14** |

The performance of different models on CIFAR-10 and CIFAR-100 is shown in Table 1. The following observations can be made from it:

1) Compared with PreActResNet, TM-ResNet constructed using the proposed method achieves higher test set accuracy and lower test set loss. This indicates that TM-ResNet has better performance than ResNet before the improvement.

2) In the experiments, comparisons were made between models before and after the improvement on different datasets. The results show that TM-ResNet generally achieves higher accuracy with fewer parameters. This indicates that the model improved by the Taylor multi-step method proposed in this chapter (TM-ResNet) has excellent performance.

## V. CONCLUSIONS

This paper begins by elucidating the connection between numerical methods and the ResNet network architecture. It then constructs a higher-precision multi-step numerical difference method and uses this method to improve the stacking scheme of ResNet. The improved model has two main features: 1) The enhanced model is supported by the theory of Ordinary Differential Equations (ODEs), which allows it to be extended to the vast majority of neural networks with residual connections. Additional experimental results in this paper also confirm this point; 2) The improvement method is simple and easy to implement. It only requires modifications to the residual connection structure and can enhance the model's performance while reducing the number of parameters. Subsequently, this chapter thoroughly validates the superiority of the proposed stacking scheme through extensive experiments, and the results demonstrate the potential of reconstructing ResNet-like networks through higher-order numerical methods. Compared to empirically designed convolutional neural network improvement methods.